# Evaluating Privacy-Preserving Machine Learning in Critical Infrastructures: A Case Study on Time-Series Classification

Dominique Mercier, Adriano Lucieri, Mohsin Munir, Andreas Dengel, and Sheraz Ahmed

*Abstract*—With the advent of machine learning in applications of critical infrastructure such as healthcare and energy, privacy is a growing concern in the minds of stakeholders. It is pivotal to ensure that neither the model nor the data can be used to extract sensitive information used by attackers against individuals or to harm whole societies through the exploitation of critical infrastructure. The applicability of machine learning in these domains is mostly limited due to a lack of trust regarding the transparency and the privacy constraints. Various safety-critical use cases (mostly relying on time-series data) are currently underrepresented in privacy-related considerations. By evaluating several privacy-preserving methods regarding their applicability on time-series data, we validated the inefficacy of encryption for deep learning, the strong dataset dependence of differential privacy, and the broad applicability of federated methods.

*Index Terms*—Time-Series Classification, Privacy-Preserving Machine Learning, Differential Privacy, Federated Learning, Secure Sharing, Critical Infrastructure, explainable AI

## I. Introduction

THE growing involvement of automated decision making in safety-critical areas such as healthcare, transportation, automation, and infrastructure management poses special challenges for the protection of societies and individuals. While flourishing areas like eXplainable AI (XAI), which deals with the verification of system functions and building of stakeholders' trust, offer first practical solutions for real-world applications, others are yet left mostly unconsidered by the broader community. Besides the right to an explanation, as defined in the GDPR [1], providers and developers of modern AI systems are furthermore legally bound to ensure confidentiality of user data and to implement the right to be forgotten. The emerging field of Privacy-Preserving Machine Learning (PPML) tackles problems arising through the disclosure and transfer of sensible information during training and inference of machine learning models [2]. Potential privacy breaches are multidimensional, including training data, input, output, and model privacy addressed by different concepts such as Federated Learning (FL), Secure Multiparty Computation (SMPC), Differential Privacy (DP), and Homomorphic Encryption (HE). As different privacy dimensions are often only partially covered by single PPML methods, sophisticated privacy concepts consisting of diverse measures must be developed to ensure safe deployment in areas of critical infrastructure.

The targeted reconstruction of training samples from a model's weights [3] is just one of many possible scenarios [4], [5] that could lead to unintentional data leakage in the real-world deployment of data-driven algorithms. Such privacy breaches could lead to more severe consequences beyond the theft of sensible user information or companies' intellectual property. Disclosure of vulnerabilities in critical infrastructures like energy grids, hospitals, and transportation systems through digital data processing could be exploited by attackers, potentially harming entire societies.

The goal pursued in PPML is sometimes opposed to motivations of XAI research [4], [6] and computationally efficient machine learning. It is therefore especially important to emphasize the early development of advanced PPML techniques allowing sufficient explanation and computational efficiency.

First frameworks for PPML have emerged in recent years [7], [8]. Despite the variety of developed methods, reviews of PPML methods usually focus on the image domain [9], [10] while the specific applicability of methods on time-series data is usually left unattended. Sequential time-series data poses different challenges compared to visual data and plays an enormous role in many critical applications such as energy infrastructure, healthcare, and automation and must therefore be examined in isolation.

This work evaluates the transferability of recent state-of-the-art PPML methods to the time-series domain and reports the lessons learned.
Our contributions include:

- Extensive performance benchmarking considering different privacy preserving methods and architectures in the time-series highlighting their applicability and limitations.
- Detailed influence analysis of hyper-parameters on DP and FL.
- A fusion approach combining DP and FL resulting in minor accuracy loss and increased privacy.
- Evaluation of runtime and performance drop in encrypted Secret Sharing training and inference, respectively highlighting the significant overhead of HE and limited applications areas.

Manuscript created April 16, 2021; revised August 7, 2021 and October 8, 2021; accepted October 18, 2021. Date of publication XX X, 2021. This work was supported by the BMBF projects SensAI (BMBF Grant 01IW20007) and the ExplAINN (BMBF Grant 01IS19074). *(D. Mercier and A. Lucieri are co-first authors.) (Corresponding authors: D. Mercier, A. Lucieri.)*

D. Mercier, A. Lucieri, M. Munir, and A. Dengel are with the German Research Center for Artificial Intelligence (DFKI) GmbH, 67663 Kaiserslautern, Germany, and Technical University Kaiserslautern, 67663 Kaiserslautern, Germany (e-mail: firstname.lastname@dfki.de).

S. Ahmed is with the German Research Center for Artificial Intelligence (DFKI) GmbH (e-mail: sheraz.ahmed@dfki.de).

## II. RELATED WORK

Recent achievements in the XAI domain and major efforts towards the explanation of algorithmic decisions [11] led to impressive achievements, including the reconstruction of input from trained models. Research has shown that recovering sensitive data from models is possible and results in a strong need for methods assuring privacy. Carlini et al. [5] have shown that it is possible to reconstruct a network's training data in the natural language domain and measure the likelihood of a model to reconstruct textual features. Coavoux et al. [4] discuss the amount of information exposed by the latent representation of data leaked through insecure channels.

Several methods evolved in the last years to fulfill privacy standards required to use machine learning in critical infrastructure domains resolving weak points of plain models. Al-Rubaie and Chang [2] described different privacy threats and corresponding solutions when dealing with sensitive data. Detailed explanations of the privacy-preserving methods are provided in Section III.

Over several years different strategies evolved to understand and attack machine learning models. The model reconstruction introduced by Milli et al. [12] allows the reconstruction of rare private data through attacks on a model's latent space. This can result in data leakage, especially for modern Deep Learning (DL) methods in combination with XAI, as a model's gradients and latent information is often exposed to the public. Another attack proposed by Frederikson et al. [3] called model inversion allows the reconstruction of training data utilizing public feature vectors. Membership inference attacks were introduced by Shokri et al. [13] and allow to infer whether specific samples were part of a model's training dataset. Anonymization was a common approach to protect sensitive data from providing individual information to third parties. However, methods were developed which allow inverting this process, rendering sole anonymization useless for the protection of sensitive information. For a more detailed report on data anonymization, we refer to Saranya et al. [14], providing an overview of the possibilities to anonymize data and problems arising from this approach.

The exponential increase of digital data produced by diverse parties led to interactions between different digital stakeholders worldwide. The field of collaborative data analysis has therefore increasingly moved to the industry domain and research, leading to additional challenges concerning data privacy. Zhang et al. [15] describe challenges and problems arising through this distributed setting. The authors propose a solution that fulfills cryptographic and distributed requirements to establish a secure environment. Besides such approaches, perturbation of private data aiming to blur individual information while retaining important population statistics is one of the most powerful solutions to achieve privacy for machine learning. Ji et al. [16] analyzed several aspects of differential privacy methods including a review and the applicability of these methods to different models.

In the privacy-preserving time-series domain Imtiaz et al. [17] leveraged the benefits of data feature clustering to enhance distributed private training, leading to increased robustness of results for federated settings. Moreover, Zheng et al. [18] explored a similar direction, utilizing tree-based methods to collect data similarities, creating a query scheme. Yue et al. [19] performed private medical sequence data analysis using fully encrypted LSTM networks. Erdemir et al. [20] investigated private sharing of time-series data for reinforcement learning using mutual information to measure the information revealed by the system.

The survey work of Zhang et al. [21] presents a wide variety of existing privacy methods, their computation overhead and limitations. Tanuwidjaja et al. [10], [22] surveyed different privacy-preserving approaches as well as important terminology related to PPML.

As opposed to the development of specific privacy-preserving methods for the time-series domain, there is still a lack of extensive evaluation of readily available privacy-preserving methods, commonly used in other domains. In this work, we attempt to benchmark the most common PPML techniques and open-source frameworks regarding their applicability on time-series data.

## III. EVALUATED METHODS

### A. Differential Privacy

The field of DP deals with the maximization of population-level information extracted from sensitive data while minimizing the probability of extracting information about individual samples or data subsets. The method evaluated in this work is the Differentially Private SGD algorithm (DP-SGD) [23]. DP-SGD engages in the optimization of the DL model and ensures training data privacy in the final model weights by clipping, averaging, and adding noise to the gradients over different subsets of samples. Other DP methods aim at perturbing the input [24], output or optimization objective [25] of the model.

We use the DP-SGD implementation provided by Tensorflow Privacy[1]. The influence of DP is investigated under variation of the batch size, gradient clipping threshold, and noise level.

### B. Federated Learning

Federated Learning describes the idea of distributing the optimization of a machine learning algorithm to multiple remote clients, allowing them to securely contribute their private share of data locally to the model training. The term was first coined in 2015 by McMahan et al. [26] introducing the FedAVG algorithm, consists of a central server retrieving model gradients from each client after every iteration. An averaged model is redistributed to all clients at the beginning of the next iteration.

The applicability of the widely used FedAVG method is evaluated under variation of different training parameters including the batch size, number of remote clients, number of contributing clients per epoch, and stratification of the data. Experiments are conducted using the implementations from Tensorflow Federated[2].

---
[1] https://github.com/tensorflow/privacy
[2] https://www.tensorflow.org/federated

TABLE I
UEA & UCR DATASETS RELATED TO CRITICAL INFRASTRUCTURE.

| Sector & Dataset | Train | Test | Length | Chls. | Classes |
|---|---|---|---|---|---|
| **Communications** | | | | | |
| UWaveGestureLibraryAll | 896 | 3582 | 945 | 1 | 8 |
| **Critical manufacturing** | | | | | |
| FordA | 3601 | 1320 | 500 | 1 | 2 |
| **Energy** | | | | | |
| ElectricDevices | 8926 | 7711 | 96 | 1 | 7 |
| **Food and agriculture** | | | | | |
| Crop | 7200 | 16800 | 46 | 1 | 24 |
| Strawberry | 613 | 370 | 235 | 1 | 2 |
| **Information Technology** | | | | | |
| Wafer | 1000 | 6164 | 152 | 1 | 2 |
| **Public health** | | | | | |
| ECG5000 | 500 | 4500 | 140 | 1 | 5 |
| FaceDetection | 5890 | 3524 | 62 | 144 | 2 |
| MedicalImages | 381 | 760 | 99 | 1 | 10 |
| NonInvasiveFetalECGThorax1 | 1800 | 1965 | 750 | 1 | 42 |
| PhalangesOutlinesCorrect | 1800 | 858 | 80 | 1 | 2 |
| **Telecommunications** | | | | | |
| CharacterTrajectories | 1422 | 1436 | 182 | 3 | 20 |
| HandOutlines | 1000 | 370 | 2709 | 1 | 2 |
| **Transportation systems** | | | | | |
| AsphaltPavementType | 1055 | 1056 | 1543 | 1 | 3 |
| AsphaltRegularity | 751 | 751 | 4201 | 1 | 2 |
| MelbournePedestrian | 1194 | 2439 | 24 | 1 | 10 |

Federated ensembling is conducted for comparison with FedAVG. Ensembling is a popular tool to improve the performance of multiple weak learners in machine learning [27]. If not stated otherwise, weighted softmax averaging is used as ensembling scheme.

## C. Secret Sharing & Homomorphic Encryption

Homomorphic encryption schemes allow performing computations on ciphertexts, resulting in equivalent computations as applied to their corresponding plain texts. HE still comes with a lot of limitations that impede its utility in practical Deep Learning applications [28]. Due to these persisting issues, we performed limited evaluation on HE through Secret Sharing.

The idea of Secret Sharing is to split a secret into $n$ uninformative shares distributed to $n$ independent clients. Conversions between arithmetic and binary Secret Sharing schemes used in the CrypTen [8] framework are partially homomorphic and use private addition and multiplication to allow the computation of linear, non-linear as well as comparator operations. We evaluate the feasibility of the Feature Aggregation use case where each client possesses an encrypted share of features that can be privately combined for training. This scenario can occur when different energy providers collaborate by privately sharing partial grid features or hospitals sharing partial electronic health records of the same patient. Moreover, we assess the feasibility of inferring on an encrypted version of a publicly trained model.

## IV. DATASETS

We selected a subset of datasets from UEA & UCR [29] repositories for our experimentation, addressing privacy critical classification tasks from some of the most critical sectors to benchmark the applicability and performance of existing privacy-preserving methods on sensitive time-series data. The datasets cover high-stakes fields such as energy, communication, transportation, industry, and healthcare. In addition to the variety of tasks, sequence lengths, numbers of channels,

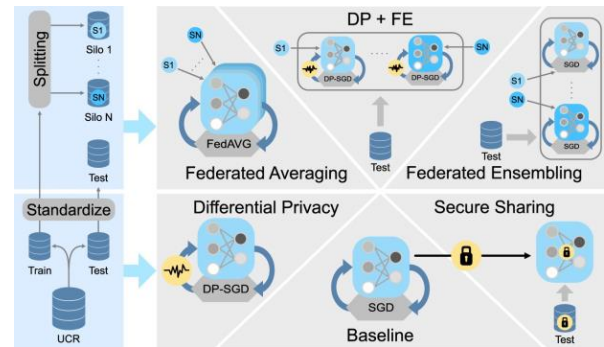

Fig. 1. **Experimental Setup.** Visualization of the different approaches, their combination and data used by those methods. DP + FE refers to the fusion approach using differential privacy and federated ensembling.

and dataset sizes, the subset addresses different types of data including sensor or EEG/ECG data.

Table I lists the different characteristics of the datasets used in this study.

## V. EXPERIMENTS & RESULTS

Figure 1 outlines the experimental setup in which different privacy-preserving data analysis methods are applied on the same preprocessed data to assure commensurability. In case of federated training, data is split into $N$ distinct data silos and experiments were conducted 5 times with different reproducible splits to account for variations in data distribution. In addition, we always selected the best run out of the 5 as representative performance for the corresponding setup.

We constructed a 1D version of AlexNet as baseline. Due to its sufficiently large amount of parameters, the network can properly generalize on the utilized datasets while still remembering parts of the training data, thus leaving room for improvement of data privacy. Every model is trained for 100 epochs using softmax cross-entropy loss with early stopping, SGD optimizer, if not stated otherwise and the learning rate is halved upon plateauing of the validation loss. Complete implementation details can be found in our repository[3].

### A. Experiment 1 - Performance Benchmarking

Preserving privacy in data analysis usually involves the disguise of sensitive information and therefore an inherent tradeoff between privacy and model performance. The missing information would have potentially contributed to solving the problem at hand, as targeted partial disguise of non-relevant information is nearly impossible in complex, high-dimensional data. In our first experiment series, we evaluated an AlexNet for all evaluated privacy-preserving methods mentioned in Section III, comparing their performance on the complete selection of datasets. This performance benchmarking does not yet provide any information about the amount of preserved privacy but serves as an initial comparison of the baseline models' performance as compared to the application of PPML methods such as DP, FL, and Secure Sharing.

[3] https://github.com/DominiqueMercier/PPML-TSA

A direct comparison of the approaches highlights a prevalent performance decrease when applying methods with higher privacy levels. However, most performance losses are in a reasonable frame which would not impede practical application. Table II shows the detailed comparison of weighted F1-scores for all evaluated methods and datasets.

*a) Differential Privacy:* Except for some datasets, comparable performance has been achieved by the DP-SGD approach. However, all datasets except for *HandOutlines* and *AsphaltRegularity* exhibit varying drops in performance. It appears that the application of DP overall results in notable performance losses for many datasets. This might indicate a sensitivity of neural networks regarding the clipping of gradients and the addition of noise. Both privacy and performance highly depend on the selection of the correct hyper-parameters. Experience showed that there seems to be no general rule, except for empirical testing, leading to suitable hyper-parameters resulting in an optimal trade-off.

*b) Federated Learning:* Overall results show a similar performance loss as compared to DP, disregarding small datasets resulting in non-converging models. Surprisingly, the simple ensemble approach has shown much better performance compared to both previous approaches. However, it has to be noted that in contrast to FedAVG and DP, ensembling does not provide any protection against model inversion or similar privacy attacks. The coexistence of multiple models trained on fewer data could even simplify such attacks in some cases.

### B. Experiment 2 - Architecture comparison

Furthermore, we evaluated the impact of the different methods on a variety of deep network architectures. We used a selection of five common DL architectures (AlexNet, LeNet FCN, FDN, LSTM) for experimentation to assure significance of our claims. It is important to notice that the models vary in their number of parameters and we do not compare across the models. We focused on each architecture in an isolated way to evaluate the impact of the privacy methods applied to them. The FCN and FDN structures are aligned with the respective parts in AlexNet. The LSTM consists of two bidirectional LSTM layers. These architectures cover the main set of layers used in time-series analysis. Moreover, we excluded transformer architectures in addition due to dataset-specific knowledge and embeddings required as well as further obstacles like model size, computational expense and lack of compatibility with the used frameworks.

In Table III we show that the average performance trade-off of AlexNet when using privacy methods is superior to the other models when applying privacy-preserving methods. In addition, we see that LeNet resulted in bad performance across almost all setups. Considering the generally lower performance of LeNet on the baseline models, it can be assumed that these issues arise from the reduced model capacity as compared to AlexNet. Furthermore, only the AlexNet and FDN network were able to converge across all the setups whereas the FCN and LSTM converged for all setups except one. However, besides LeNet all methods showed to be compatible with the most common privacy-preserving methods.

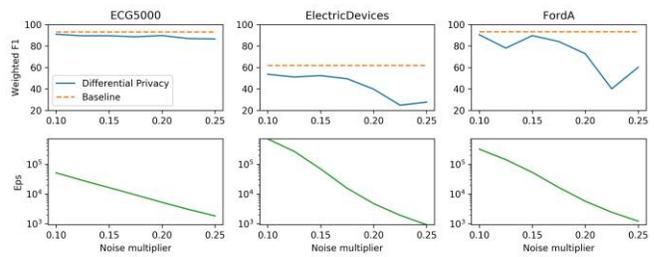

Fig. 2. **Performance vs Noise.** Evaluation of the loss in weighted F1-score and change in privacy when using different noise multipliers. Lower values of *Eps* correspond to higher privacy.

### C. Experiment 3 - Differential Privacy: Hyper-parameter Evaluation

The impact of different hyper-parameters on the privacy obtained by DP-SGD is evaluated next. We selected three representative datasets from different domains to perform evaluation, due to their varying sequence lengths, training data sizes, and the number of classes. The *FordA* dataset covers an anomaly detection task whereas the *ECG5000* and *ElectricDevices* datasets cover classification tasks. All parameters except for the noise multiplier are kept fixed as it has the most significant impact on the privacy-accuracy trade-off. Each run is performed with gradient clipping threshold set to 1.0 and a batch size of 32. Moreover, we examine the impact on the privacy level when changing each training parameter in isolation. This impact can be computed independently from model training and is therefore evaluated using a large number of different conditions.

Figure 2 provides detailed insight on the impact of noise on model performance and the corresponding change in privacy. The *Eps* value on the y-axis is an indicator for privacy. A detailed explanation of the parameter including the mathematical background can be found in [30]. In this analysis, it is enough to note that *Eps* depends on multiple different parameters and that lower values indicate higher privacy levels. The ratio of noise added to the gradients is controlled by the noise multiplier $n_E$, where the gradient is left unaltered for $n_E = 0$ but privacy is only increased for $n_E > 0$. Larger values of $n_E$ bear the risk of generating noise that dominates the actual gradient information, rendering fine-tuning crucial.

Our results show that for all datasets the performance decreases significantly after a certain value of $n_E$. *ECG5000* exhibits a relatively low and linear decrease of 3% when changing $n_E$ from 0.1 to 0.25. This does not hold for the remaining datasets. *ElectricDevices* has a stable F1-score up to $n_E = 0.175$ but then drops significantly. Similar behavior as exhibited by *FordA*. Moreover, *FordA* covers a binary anomaly detection task that reflects an unacceptable performance loss for noise multiplier values larger than 0.2.

The results can be summarised as follows: The *Eps* value is a good and inexpensive indicator that can be used to provide a solid estimate of the privacy achieved in a specific parameter setup, prior to model training. However, its absolute value is difficult to interpret and greatly depends on the dataset. The noise multiplier $n_E$ has a drastic impact on the model performance but this impact is dependent on the data distribution and problem at hand.

TABLE II
**PERFORMANCE BENCHMARKING.** COMPARISON OF BASELINE ALEXNET MODEL AND DIFFERENT PRIVACY-PRESERVING METHODS REPORTING BEST WEIGHTED F1-SCORES. *N* CORRESPONDS TO THE NUMBER OF CLIENTS USED IN FEDERATED SETTINGS. RESULTS OF NON-CONVERGING MODELS ARE STRUCK OUT.

| Dataset | Baseline | Diff. Privacy | FedAVG N=2 | FedAVG N=4 | Fed. Ens. N=2 | Fed. Ens. N=4 |
|---|---|---|---|---|---|---|
| AsphaltPavementType | 88.30 | 81.90 | 85.22 | 80.88 | 88.93 | 86.22 |
| AsphaltRegularity | 98.93 | 98.93 | 98.54 | 96.27 | 99.07 | 98.80 |
| CharacterTrajectories | 99.37 | 97.88 | 96.98 | 89.29 | 99.09 | 98.74 |
| Crop | 75.16 | 48.70 | 56.64 | 38.06 | 74.18 | 72.14 |
| ECG5000 | 93.37 | 89.58 | 88.57 | 87.49 | 93.33 | 92.70 |
| ElectricDevices | 64.01 | 52.71 | 65.14 | 64.76 | 65.91 | 65.90 |
| FaceDetection | 63.58 | 51.43 | 62.11 | 62.34 | 64.55 | 64.59 |
| FordA | 92.80 | 91.06 | 90.90 | 85.91 | 93.49 | 93.11 |
| HandOutlines | 91.29 | 98.81 | 86.99 | 85.73 | 91.31 | 89.85 |
| Medical Images | 77.20 | 51.21 | ~~34.95~~ | ~~34.95~~ | 72.14 | 64.13 |
| MelbournePedestrian | 94.94 | 86.55 | 18.44 | 20.77 | 86.19 | 87.60 |
| NonInvasiveFetalECGThorax1 | 90.81 | 78.01 | 35.60 | ~~5.15~~ | 90.65 | 87.36 |
| PhalangesOutlinesCorrect | 82.29 | ~~46.60~~ | ~~46.60~~ | ~~46.60~~ | 79.97 | 78.91 |
| Strawberry | 96.77 | ~~50.36~~ | ~~50.36~~ | ~~50.36~~ | 95.43 | 95.41 |
| UWaveGestureLibraryAll | 96.06 | 90.26 | 92.33 | 89.91 | 95.54 | 93.52 |
| Wafer | 99.50 | 98.10 | ~~84.12~~ | ~~84.12~~ | 98.67 | 98.19 |
| Average | 87.77 | 75.76 | 68.34 | 63.91 | 86.78 | 85.45 |

TABLE III
**ARCHITECTURE BENCHMARKING.** COMPARISON OF DIFFERENT MODEL ARCHITECTURES REPORTING WEIGHTED F1-SCORES. *N* CORRESPONDS TO THE NUMBER OF CLIENTS USED FOR THE FEDERATED APPROACHES. RESULTS OF NON-CONVERGING MODELS ARE STRUCK OUT.

| Architecture | Dataset | Baseline | Privacy-Preserving Methods | | | Average |
|---|---|---|---|---|---|---|
| | | | Diff. Privacy | FedAVG N=4 | Fed. Ens. N=4 | |
| AlexNet | ECG5000 | 93.37 | 89.58 | 87.49 | 92.53 | |
| | ElectricDevices | 64.01 | 52.71 | 64.76 | 62.80 | |
| | FordA | 92.80 | 91.06 | 85.91 | 92.80 | |
| | **Average** | 83.39 | 77.78 | 79.39 | 82.71 | 79.96 |
| LeNet | ECG5000 | 87.70 | ~~43.03~~ | ~~43.04~~ | 43.04 | |
| | ElectricDevices | 63.28 | ~~60.18~~ | 31.10 | 61.11 | |
| | FordA | ~~31.58~~ | ~~35.11~~ | ~~35.61~~ | 35.12 | |
| | **Average** | 60.85 | 46.11 | 36.58 | 46.42 | 43.04 |
| FCN | ECG5000 | 88.43 | 88.51 | 86.91 | 91.67 | |
| | ElectricDevices | 50.05 | 46.16 | ~~9.46~~ | 60.11 | |
| | FordA | 69.70 | 84.38 | 61.86 | 91.82 | |
| | **Average** | 69.39 | 73.02 | 52.74 | 81.20 | 68.99 |
| FDN | ECG5000 | 93.08 | 88.24 | 89.89 | 90.61 | |
| | ElectricDevices | 51.50 | 53.56 | 52.89 | 52.63 | |
| | FordA | 82.58 | 67.01 | 80.30 | 76.52 | |
| | **Average** | 75.72 | 69.60 | 74.36 | 73.25 | 72.41 |
| LSTM | ECG5000 | 92.57 | 85.38 | 85.98 | 89.10 | |
| | ElectricDevices | 70.33 | 62.18 | 57.30 | 62.12 | |
| | FordA | 42.22 | ~~0~~ | 42.34 | 48.03 | |
| | **Average** | 68.37 | 49.19 | 61.87 | 66.42 | 59.16 |

Another important aspect when applying DP is the impact of other parameters such as dataset size, batch size, and the number of epochs. Using the estimation approach mentioned above, we calculated the expected *Eps* values in a controlled environment. We started with a fixed setup using 5000 samples, 100 epochs, batch size 32 and $n_E$: 0.5.

Only one of the parameters is changed at a time to assess the impact of parameters independently. The results are presented in Figure 3. The baseline is marked with vertical orange lines. Confirming intuition, the dataset size, and the noise multiplier lower increase privacy whereas the batch size and the number of epochs decrease it.

The results emphasize that the method can give a good idea about the possible setup required to achieve a certain level of privacy before training. However, the consideration of *Eps* does not provide any information about the convergence guarantees, which must be adjusted through batch size and epochs.

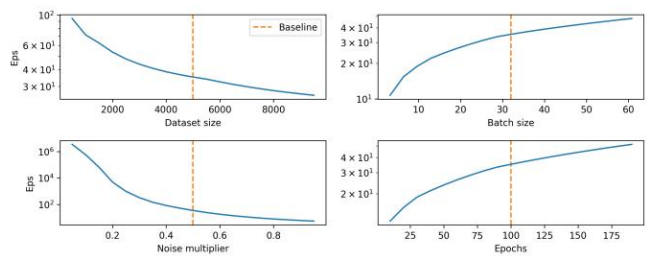

Fig. 3. **Parameter impact.** Evaluation of different parameters with respect to the privacy. Lower *Eps* values correspond to higher privacy.

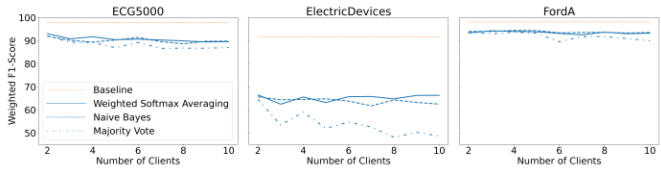

Fig. 4. **Federated Ensemble Baseline.** Performance evaluation of three different ensembling voting techniques. Weighted F1-Scores are presented for three different datasets.

### D. Experiment 4 - Federated Ensemble: Ensemble Size Evaluation

The number of participating clients, as well as the amount and quality of data contributed by individual clients, are the most critical factors in federated learning. This experiment investigates the impact of increasing numbers of clients in the most simplistic case of federated ensembling. Both batch size $b = \{8, 16, 32, 64\}$ and learning rates $lr = \{1e-2, 1e-3, 1e-4\}$ were tuned to obtain the best performance in each setting. Federated experiments are conducted 5 times to account for variations in data distribution of single data silos.

Three ensemble methods have been evaluated on the federated training of *ECG5000*, *ElectricDevices* and *FordA* datasets. Figure 4 gives an overview of the performances achieved. The results show that Weighted Softmax Averaging and Naive Bayes classification achieve similar performance on the test datasets, while ensembling by Majority Vote resulted in the worst weighted F1-scores. It can be observed that the performance of Majority Voting follows a downward trend with an increasing number of clients.

Ensembles trained on *ECG5000* and *FordA* both suffer from a minor drop in classification and anomaly detection performance whereas the F1-score for *ElectricDevices* significantly decreases with a higher number of clients.

### E. Experiment 5 - Differential Privacy in a Federated Setting

We examine the possibility to train local data in a federated setting using DP-SGD at each client machine and theoretically consider the resulting gain in privacy. Evaluation is done on all datasets for different number of clients $N = \{2, 4\}$ and batch sizes $b = \{16, 32\}$, with fixed gradient clipping parameter $L2 = 0.5$ and noise multiplier $n_E = 0.1$.

Table IV shows the results of combined differential private training of federated ensembles on all datasets. Many datasets show decent performance losses over all tested settings. Overall the results show that depending on the dataset at hand, a combination of DP and federated ensembling can be feasible to combine its strengths. Higher performance could be achieved by extensive hyper-parameter tuning on the specific use case, as experience showed that specially DP-SGD is sensitive to certain hyper-parameters.

A combination of differentially private with federated training results in a non-linear combination of the privacy levels as the privacy achieved by DP-SGD depends on the dataset size, batch size, and the number of epochs, which might vary when switching from an aggregated to a federated setting. Training in a federated setting aids training data privacy in two ways, by ensuring that a client's data remains on-site and by introducing an averaging which mitigates some model inversion attacks. The additional application of differentially private training on-site adds further noise to the process which consequentially results in an overall improvement of the training data privacy. Whether a combination of DP and FL is suitable highly depends on the dataset sizes available at individual client locations as well as the complexity of the problem and must therefore be decided on a case-by-case basis. As previously concluded in the hyper-parameter evaluation of DP, a lower dataset size, as well as a higher number of epochs, decrease privacy. Both of which are likely to be the consequence of switching from an aggregated to a distributed setting. A securely aggregated, differentially private training therefore might result in a higher privacy level as compared to local, federated training on smaller datasets.

### F. Experiment 6 - Secret Sharing Runtime Evaluation

Training and validation runtimes are major considerations for the practical applicability of data-driven methods, especially in time-critical real-time applications. We evaluate the feasibility of applying Secret Sharing to time-series applications by assessing training and validation runtimes, comparing the implementations of the same 2D AlexNet for time-series in vanilla PyTorch versus CrypTen. CrypTen has been evaluated in the most basic setting performing encrypted training with only a single client. Note that a 2D model was chosen to have a comparable number of parameters in both settings. Unlike CrypTen, vanilla PyTorch is not restricted to 2D architectures, which results in a minor slow down.

An evaluation of training and inference runtimes comparing the implementations of the same 1D AlexNet for time-series in vanilla PyTorch versus CrypTen gives a first estimate about the feasibility of encrypted Secret Sharing in practice. Table V shows that both training and inference using CrypTen is significantly slower than vanilla PyTorch in the case of CPU (roughly factor 350) and even more in the more realistic case of GPU computation. This highlights the impracticality of encrypted Secret Sharing for current real-world applications.

### G. Experiment 7 - Encrypted Inference Evaluation

In a final experiment, a different Secret Sharing scenario is considered where a model is trained on public data and encrypted for inference on secret data. We assess potential performance deviations arising from the encrypted evaluation of data and model at inference time.

Table VI shows the weighted F1-Scores obtained by private prediction on an encrypted AlexNet, trained on public data. It can be observed that the performance of *ECG5000* and *ElectricDevices* decreased negligibly and *FordA* even increased slightly. This minor deviation of the original results is expected, as encrypted computation results in some change due to noisy encryption.

TABLE IV
**Differential + Federated Ensemble.** Comparison of baseline weighted accuracies using both methods separately and their combination to achieve better privacy reporting weighted F1-scores. $N$ corresponds to the number of clients used for the federated approaches. Results of non-converging models are struck out.

| Dataset | Diff. Privacy | N=2 Fed. Ens. | N=2 DP + Fed Ens. | N=4 Fed. Ens. | N=4 DP + Fed Ens. |
|---|---|---|---|---|---|
| AsphaltPavementType | 81.90 | 88.93 | 78.44 | 86.22 | 77.96 |
| AsphaltRegularity | 98.93 | 99.07 | 97.87 | 98.80 | 96.40 |
| CharacterTrajectories | 97.88 | 99.09 | 97.72 | 98.74 | 97.66 |
| Crop | 48.70 | 74.18 | 63.18 | 72.14 | 62.99 |
| ECG5000 | 89.58 | 93.33 | 90.01 | 92.70 | 89.52 |
| ElectricDevices | 52.71 | 65.91 | 61.22 | 65.90 | 55.39 |
| FaceDetection | 51.43 | 64.55 | 51.66 | 64.59 | 51.84 |
| FordA | 91.06 | 93.49 | 93.33 | 93.11 | 91.36 |
| HandOutlines | 98.81 | 91.31 | 68.33 | 89.85 | 87.40 |
| Medical Images | 51.21 | 72.14 | 47.53 | 64.13 | 37.95 |
| MelbournePedestrian | 86.55 | 86.19 | 87.95 | 87.60 | 86.47 |
| NonInvasiveFetalECGThorax1 | 78.01 | 90.65 | 76.18 | ~~87.36~~ | ~~1.79~~ |
| PhalangesOutlinesCorrect | ~~46.60~~ | 79.97 | 62.29 | 78.91 | 50.70 |
| Strawberry | ~~50.36~~ | ~~95.43~~ | ~~50.36~~ | ~~95.41~~ | ~~50.36~~ |
| UWaveGestureLibraryAll | 90.26 | 95.54 | 93.55 | 93.52 | 92.24 |
| Wafer | 98.10 | 98.67 | 96.16 | 98.19 | 95.48 |
| Average | 75.76 | 86.78 | 75.98 | 85.45 | 70.34 |

TABLE V
**Runtime Evaluation.** Evaluation of runtimes over one batch of size 8. All Values given in seconds. Used hardware: Intel Xeon (Quad Core), Nvidia GTX 1080 Ti, 64 GB memory.

| Dataset | Framework | Training Avg (s) | Training Std (s) | Inference Avg (s) | Inference Std (s) |
|---|---|---|---|---|---|
| ECG5000 | CrypTen | 35.132 | 0.594 | 8.561 | 0.363 |
|  | PyTorch CPU | 0.105 | 0.019 | 0.024 | 0.002 |
|  | PyTorch GPU | 0.004 | 0.001 | 0.001 | 0.000 |
| ElectricDevices | CrypTen | 30.196 | 0.145 | 7.113 | 0.030 |
|  | PyTorch CPU | 0.086 | 0.005 | 0.019 | 0.000 |
|  | PyTorch GPU | 0.004 | 0.001 | 0.001 | 0.000 |
| FordA | CrypTen | 68.110 | 0.484 | 18.673 | 0.931 |
|  | PyTorch CPU | 0.186 | 0.016 | 0.050 | 0.005 |
|  | PyTorch GPU | 0.004 | 0.001 | 0.001 | 0.000 |

TABLE VI
**Encrypted Inference.** Performance loss for encrypted inference compared to baseline AlexNet reporting weighted F1-scores.

| Model | ECG5000 | ElectricDevices | FordA |
|---|---|---|---|
| AlexNet Baseline | 93.37 | 64.01 | 92.80 |
| AlexNet Enc. | 90.10 | 63.14 | 93.03 |

## VI. Discussion

The image domain is usually in the focus of new ML developments due to the ease of problem understanding and intuitive interpretation of context. The conducted experiments serve as a first overview of the applicability and usability of current state-of-the-art PPML applications for time-series classification in safety-critical domains. Our experience with available open-source frameworks showed that PPML methods applicable to time-series classification already exist. However, for some applications minor and sometimes major adjustments are required for the proper utilization as most of the frameworks are not in a productive state and offer only limited support concerning features specifically required for time-series. For instance, most of the frameworks cover only implementations for 2D image processing although time-series classification is a very important modality that is used in almost all of the sixteen safety-critical domains.

During experimentation, some challenges of PPML specific to the domain of time-series classification were revealed. DP is a useful tool for ensuring the privacy of remote time-series data. The applicability of the method is however strongly linked with a trade-off between privacy and accuracy, which depends a lot on the dataset and machine learning task at hand. The selection of the right hyper-parameters to ideally balance this trade-off is especially complicated in real-world scenarios, where model providers have to select hyper-parameters for unseen data on the client-side. In such cases, we would recommend a top-down strategy in which the noise and gradient clipping parameters should initially provide maximum privacy in critical infrastructure use cases while gradually being relaxed until an acceptable model performance is achieved while data privacy is still tolerable. However, a set of possible setups can be discovered using the mathematical equation to compute the privacy value related to the differential privacy approach. Doing so provides a possible set. It is not possible to know the degree of network convergence without training the network using the actual setting. Summarizing our findings, it is highly beneficial to know the dataset features and their susceptibility with respect to noise. Therefore, the understanding of the classification task and the value ranges can be used to approximate suitable parameters for the approach.

Both FL and Secret Sharing did not prove to present unusual challenges when applied to time-series classification. For FL in general, but especially in time-series classification, it is of prime importance that the preprocessing of data is performed identically. Whereas preprocessing of other data types such as images is much more natural and standardized, preprocessing of time-series data is very application and problem dependent and must be communicated to all participating clients in a learning federation. The application of time-series data partially alleviates the common downside of the high temporal and computational cost related to homomorphic or partially homomorphic encrypted computation. Despite HE exhibiting unbearable computation times, making it unfeasible for practi-cal application in critical infrastructure, the private sharing of data for encrypted inference proved to be a suitable approach. Furthermore, our experiments using federated learning showed that the combination of privacy-persevering methods, namely DP and FL, performs similarly well. This indicates that the combination of several feasible privacy-preserving methods can be used to develop a comprehensive privacy concept for real-world applications. Overall the performance of FL is comparable to the DP approach. Whereas DP is more sensitive to hyper-parameters like noise, FL is more sensitive towards small dataset sizes and uneven data distributions. Intuitively, the combination of both approaches suffers from both aspects and achieved a lower average accuracy but an increase in privacy. However, if certain aspects of the datasets are known it is possible to adjust for these aspects.

## VII. CONCLUSION

Together with XAI, PPML is a key technology paving the way towards the omnipresent application of AI in critical infrastructure systems by allowing to leverage synergies from the collaboration of multiple private entities. This safe collaboration has the potential to achieve safe and methodically transparent deployment of high-performing data-driven algorithms in critical real-world high-stakes decision scenarios. We benchmarked methods and open-source frameworks to provide a first overview of the applicability of PPML methods to the time-series domain, which plays a crucial role in a variety of critical infrastructure application fields like energy, industry, and healthcare, and highlighted challenges specific to this particular type of input data. Our benchmarking covers different model architectures commonly used in the time-series domain. Furthermore, we used a set of carefully selected datasets that cover various different aspects with respect to their domain, data shape and task. Our findings highlight that it is possible to successfully apply DP, FL, and our fusion approach to different architectures and datasets. Furthermore, our findings highlight the importance of a proper hyper-parameter selection for the DP and the drawbacks using HE with respect to the computational effort. For future research, research communities need to engage in the joint development of explainable and privacy-preserving ML solutions to balance their competing objectives and achieve a broad applicability in industrial use cases. Initially, efforts should be spent on assessing the compatibility of different PPML methods with existing XAI techniques. Moreover, computational and communication overhead still constitute a major barrier to the practical applicability of some PPML techniques e.g. HE.

## ACKNOWLEDGEMENTS

We thank all members of the Deep Learning Competence Center at the DFKI for their comments and support.

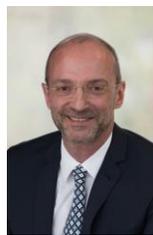

**Andreas Dengel** is Scientific Director at DFKI GmbH in Kaiserslautern. In 1993, he became Professor in Computer Science at TUK where he holds the chair Knowledge-Based Systems. Since 2009 he is appointed Professor (Kyakuin) in Department of Computer Science and Information Systems at Osaka Prefecture University. He received his Diploma in CS from TUK and his PhD from University of Stuttgart. He also worked at IBM, Siemens, and Xerox Parc. Andreas is member of several international advisory boards, has chaired major international conferences, and founded several successful start-up companies. He is co-editor of international computer science journals and has written or edited 12 books. He is author of more than 300 peer-reviewed scientific publications and supervised more than 170 PhD and master theses. Andreas is an IAPR Fellow and received many prominent international awards. His main scientific emphasis is in the areas of Pattern Recognition, Document Understanding, Information Retrieval, Multimedia Mining, Semantic Technologies, and Social Media.

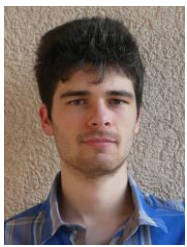

**Dominique Mercier** received his Master degree in computer science from the Technische Universitaet Kaiserslautern, Germany in 2018. The topic of his Master thesis was 'Towards Understanding Deep Networks for Time Series Analysis'. Currently, he is pursuing his Ph.D. at German Research Center for Artificial Intelligence (DFKI GmbH) under the supervision of Prof. Dr. Prof. h.c. Andreas Dengel. His areas of interests include the interpretability of deep learning methods, time series analysis and document analysis. His work includes the development of novel interpretability methods for deep neural networks for time-series analysis. Furthermore, he actively working in the NLP domain with e focus to citation and community management.

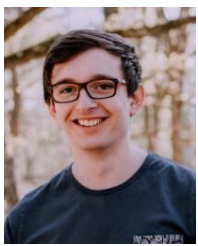

**Adriano Lucieri** completed his BE in Mechatronic Engineering from Duale Hochschule Baden-Württemberg (DHBW) Mannheim and MS in Mechatronic Systems Engineering from Hochschule Pforzheim in Germany. He is presently pursuing PhD from Technische Universität Kaiserslautern (TUK), Germany and is also working as Research Assistant at Deutsches Forschungszentrum für Künstliche Intelligenz GmbH (DFKI). His research focus lies on improving the explainability and transparency of Computer-Aided Diagnosis (CAD) systems based on Deep Learning for medical image analysis. His work includes concept-based explanation of skin lesion classifiers as well as the localization of concept regions in input images.

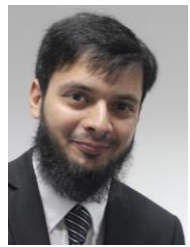

**Sheraz Ahmed** is Senior Researcher at DFKI GmbH in Kaiserslautern, where he is leading the area of Time Series Analysis and Life Science. He received his MS and PhD degrees in Computer Science from TUK, Germany under the supervision of Prof. Dr. Prof. h.c. Andreas Dengel and Prof. Dr. habil. Marcus Liwicki. His PhD topic is Generic Methods for Information Segmentation in Document Images. Over the last few years, he has primarily worked on development of various systems for information segmentation in document images. His research interests include document understanding, generic segmentation framework for documents, pattern recognition, anomaly detection, Gene analysis, medical image analysis, and natural language processing. He has more than 80 publications on the said and related topics including three journal papers and two book chapters. He is a frequent reviewer of various journals and conferences including Pattern Recognition Letters, Neural Computing and Applications, IJDAR, ICDAR, ICFHR, and DAS.

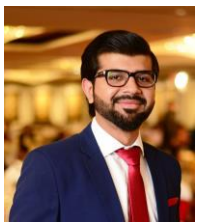

**Mohsin Munir** received his Masters degree in computer science from the University of Kaiserslautern, Germany. He did internships at RICOH (Japan) and BOSCH (Germany) during his Master's degree. The topic of his Masters thesis was 'Connected Heating System's Fault Detection using Data Anomalies and Trends'. Currently, he is pursuing his Ph.D. in Computer Science at German Research Center for Artificial Intelligence (DFKI GmbH) under the supervision of Prof. Dr. Prof. h.c. Andreas Dengel. His research topic is 'Time Series Forecasting and Anomaly Detection'. His research interests are time series analysis, deep neural networks, forecasting, predictive analytics, and anomaly detection. During his Ph.D., he did a research internship at Kyushu University (Japan) under the supervision of Prof. Seiichi Uchida.